\documentclass{article}

\usepackage[table,HTML,dvipsnames]{xcolor}
\usepackage[preprint]{corl_2026} 

\usepackage{graphicx}
\usepackage{amsmath}
\usepackage{amssymb}
\usepackage{booktabs}
\usepackage{multirow}
\usepackage{enumitem}
\usepackage{tabularx}
\usepackage{siunitx}
\usepackage{makecell}
\usepackage{multirow}
\usepackage{pgfplots}
\usepackage{caption}
\usepackage{graphicx}
\usepackage{booktabs}
\usepackage{wrapfig}

\pgfplotsset{compat=1.18}

\definecolor{tablehead}{HTML}{F4F6F8}
\definecolor{tablerow}{HTML}{FAFAFA}

\newcommand{\methodname}{{NavWAM}}

\title{NavWAM: A Navigation World Action Model for Goal-Conditioned Visual Navigation}

 \author{
    Daichi Azuma$^{1}$\quad Taiki Miyanishi$^{1}$\quad Koya
  Sakamoto$^{1}$\quad Shuhei Kurita$^{2}$\quad Yaonan Zhu$^{1}$ \\
    \bfseries Petr Khrapchenkov$^{3}$\quad Motoaki Kawanabe$^{4}$\quad Yusuke
  Iwasawa$^{1}$\quad Yutaka Matsuo$^{1}$ \\[4pt]
    $^{1}$The University of Tokyo\quad $^{2}$National Institute of
  Informatics\quad
    $^{3}$AIRoA\quad $^{4}$ATR \\[2pt]
    \texttt{daichi.azuma@weblab.t.u-tokyo.ac.jp}
}

\begin{document}
\maketitle

~

\begin{abstract}
Goal-conditioned visual navigation requires a robot to act under partial observability by anticipating how its motion will change the future egocentric view and whether that change brings it closer to the goal.
Navigation world models provide such visual foresight, but they remain prediction modules that require an external planner to convert predicted futures into closed-loop control.
We propose Navigation World Action Model (NavWAM), a diffusion-transformer policy that turns navigation world-model prediction into executable action by representing future observations, goal-progress values, and action chunks in a shared latent sequence.
By learning future prediction jointly with the action and value targets that determine closed-loop behavior, NavWAM makes visual foresight directly usable for robot control.
We build NavWAM through simulation pretraining and real-robot adaptation, and evaluate it on image-goal navigation against planning-based world models and a representative direct navigation policy.
Across offline benchmarks and closed-loop real-robot deployment, NavWAM improves over planning-based world-model baselines in our evaluations while using the default policy mode without CEM-style action search.

Project page: \url{https://dachii-azm.github.io/navwam/}
\end{abstract}


~\section{Introduction}
\label{sec:introduction}
%
Recent goal-conditioned navigation policies have made substantial progress by learning to map egocentric observations and goals directly to actions from diverse navigation data~\citep{gnm, vint, nomad, lelan, omnivla}.
These methods are strong and efficient closed-loop action predictors, but navigation under partial observability often requires more than reacting to the current view.
A robot must anticipate how its motion will change the future egocentric observation and whether that change will bring it closer to the goal.
Direct policies may acquire such foresight implicitly from action supervision, but they are not explicitly trained to predict the visual consequences of their actions.
This leaves a central robot-learning question: how can explicit visual foresight be made useful for executable closed-loop control?

\begin{figure}[t]
\begin{center}
\includegraphics[keepaspectratio, width=\textwidth]{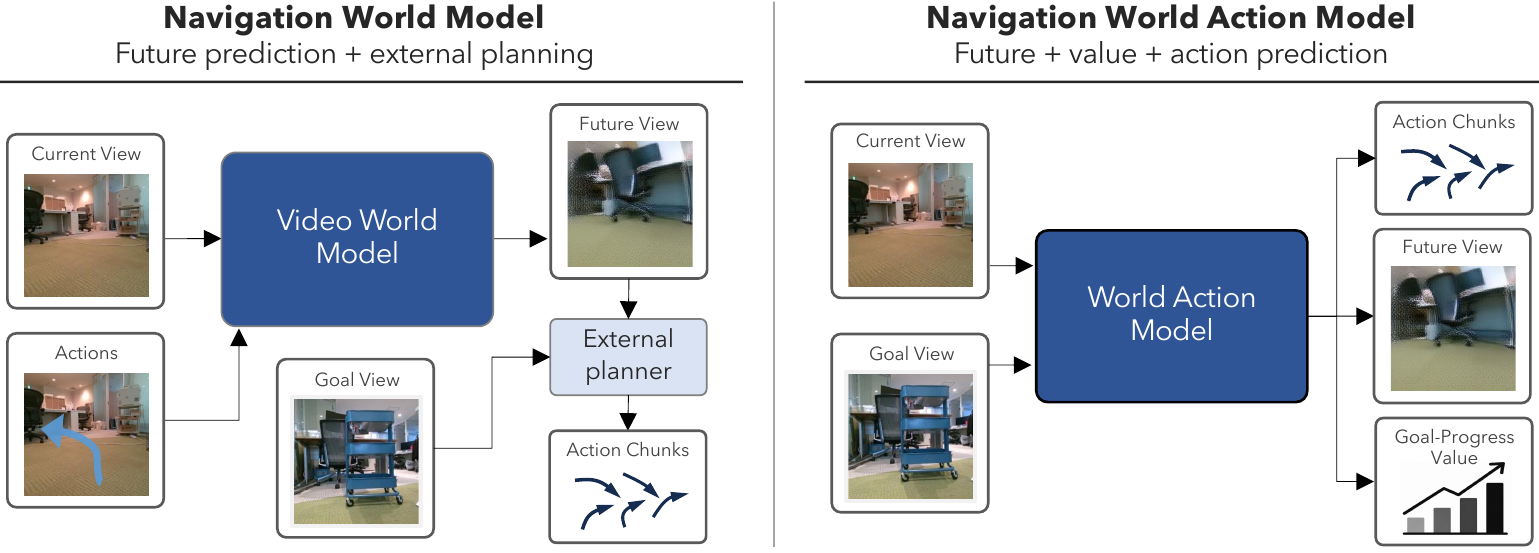}
\end{center}
\vspace{-0.3cm}
\caption{
Prior navigation world models predict future views for candidate actions and rely on external planning. \methodname{} instead predicts future egocentric views, goal-progress values, and executable action chunks within one policy representation, turning visual foresight into a closed-loop navigation policy.
}
\label{fig:teaser}
\vspace{-0.3cm}
\end{figure}


Visual prediction approaches for navigation make foresight explicit by synthesizing future egocentric observations along possible paths~\citep{pathdreamer, navdreamer, schrodingersnavigator}.
Navigation World Models (NWMs) are particularly relevant because they learn action-conditioned predictive models whose generated futures can be scored against the goal and used for planning~\citep{bar2024navigationworldmodels}.
However, NWMs remain prediction modules rather than action-producing policies.
At inference time, they rely on an external planner that searches over candidate actions based on predicted future outcomes before selecting one for execution.
This separation creates a bottleneck for robot deployment: closed-loop behavior depends on external planning choices rather than on the learned world model alone.

We propose the \textbf{Navigation World Action Model (\methodname)}, a diffusion-transformer policy that turns navigation world-model prediction into executable action.
The key idea is to place future egocentric observations, goal-progress values, and action chunks in a shared latent sequence, turning navigation into a joint denoising problem over future perception, goal progress, and executable motion.
At inference, \methodname{} conditions on the current egocentric observation and goal, directly outputs an action chunk, and executes it in a receding-horizon loop while retaining future-view prediction and value estimation as interpretable foresight.
This formulation makes visual foresight directly usable for robot control without CEM-style test-time trajectory optimization.
Figure~\ref{fig:teaser} illustrates this shift from external planning to joint world-action prediction.

Our design builds on recent efforts to repurpose video generation and video world models for robot control~\citep{unipi, gr1, cosmos_policy, cosmospredict2}.
Unlike prior video policies developed primarily for manipulation, \methodname{} targets goal-conditioned navigation, where the model must couple viewpoint-changing egocentric prediction with goal-progress estimation and local-frame action generation under partial observability.
This distinction is central to our setting: NWM-style methods can predict plausible future views, but action selection is still delegated to a separate planner.
\methodname{} instead learns future prediction, value estimation, and action generation in one policy representation.

We build \methodname{} through simulation pretraining and real-robot adaptation, following the NWM evaluation protocol, and evaluate it on image-goal navigation in both offline benchmarks and closed-loop real-robot deployment.
Our primary baseline is NWM~\cite{bar2024navigationworldmodels}, the closest planning-based world-model approach, and we also compare with OmniVLA~\cite{omnivla}, a representative direct navigation policy built on a larger 7B-parameter VLA backbone.
Across these settings, \methodname{} achieves better navigation performance than planning-based world-model baselines while avoiding CEM-style test-time trajectory optimization.
It also remains competitive with the larger direct-policy baseline using a 2B-parameter video backbone, while additionally producing future-view and value predictions.



This paper makes the following contributions:
\begin{itemize}[leftmargin=1.2em]
\item We propose \textbf{\methodname}, a navigation world action model that converts NWM-style visual foresight into an action-producing policy for goal-conditioned visual navigation.
\item We introduce a joint prediction formulation that represents future observations, goal-progress values, and executable action chunks in a shared latent sequence, allowing future prediction to directly support closed-loop action generation.
\item We show that \methodname{} improves over planning-based world-model baselines without CEM-style test-time trajectory optimization, while remaining competitive with a larger direct navigation policy and preserving interpretable future-view predictions.
\end{itemize}
~\section{Related Work}
\label{sec:related_work}
\paragraph{Goal-Conditioned Visual Navigation.}
Goal-conditioned visual navigation has been studied under a variety of goal specifications, including object categories~\citep{objectnav, semexp, cow, majumdar2022zson, gervet2023navigating}, target images~\citep{gnm, vint, nomad, go_stanford_dataset}, and natural-language instructions or questions~\citep{eqa, photorealisticeqa, sakamoto2024mapeqa, saxena2025GraphEQA, lelan, omnivla}.
Across these settings, the central challenge is to connect goal understanding with action selection.
Prior work has often addressed this by constructing explicit geometric, topological, or semantic representations and planning over them~\citep{cmp, ans, ntslam, trajectorydiffusion}.
More recent learned approaches reduce this modularity by directly predicting actions from observations, using diffusion policies~\citep{nomad}, large-scale reinforcement learning~\citep{poliformer}, or vision-language-action models~\citep{omnivla, cheng2024navila, navid, zhang2024uninavid}.
These methods are strong at direct action prediction, but future visual prediction, goal-progress estimation, and action generation are usually not learned within a single policy representation.
\methodname{} addresses this gap by integrating future prediction into the policy representation used for closed-loop navigation.

\vspace{-5pt}
\paragraph{Visual Foresight for Navigation.}
Future prediction provides a useful prior for navigation under partial observability.
Map-based approaches predict occupancy, semantics, or likely target locations beyond the current field of view~\citep{occant, l2m_active, sgm, shah2025foresightnav, peanut}, but do not model the future egocentric observations that the robot would receive after moving.
Pixel-space methods instead synthesize future observations along possible trajectories, as in PathDreamer~\citep{pathdreamer}, Schr\"odinger's Navigator~\citep{schrodingersnavigator}, NWM~\citep{bar2024navigationworldmodels}, and NavDreamer~\citep{navdreamer}.
These methods make visual foresight explicit, but still require a separate planner, value map, or scoring function to convert predicted futures into actions.
\methodname{} avoids this separation by jointly predicting action chunks, future egocentric observations, and goal-conditioned values within one policy representation.

\vspace{-5pt}
\paragraph{Video-Based Robot Policies.}
Recent robot learning methods have begun to use generative modeling as a policy representation.
Diffusion Policy established action diffusion as a strong formulation for visuomotor control~\citep{diffusion_policy}, and NoMaD adapted action-sequence diffusion to goal-conditioned navigation~\citep{nomad}.
A more closely related direction uses video prediction or generation inside the policy.
UniPi and Dreamitate generate videos as intermediate plans and then extract or track actions from them~\citep{unipi, dreamitate}, while GR-1 jointly predicts future images and robot actions for manipulation~\citep{gr1}.
Most closely related in model design, Cosmos Policy encodes actions, future states, and values as latent frames in a pretrained video model for visuomotor control and optional planning~\citep{cosmos_policy}.
\methodname{} follows this broader direction, but targets goal-conditioned navigation under partial observability.
Unlike manipulation-oriented settings, navigation requires coupling viewpoint-changing egocentric prediction with goal-progress estimation and local-frame action generation.
\methodname{} integrates these targets into one policy representation so that visual foresight directly supports closed-loop navigation rather than remaining a separate planning module.

~\section{Problem Setup and World-Action Formulation}
\label{sec:problem_overview}
Goal-conditioned visual navigation requires a robot to reach a specified goal from partial egocentric observations.
At time $t$, the robot observes an RGB image $o_t$ and receives a goal specification $g$.
In this work, we focus on image-goal navigation, where $g$ is a target image captured at the goal location.
The policy outputs an executable action chunk $a_{t:t+H-1} = (a_t,\ldots,a_{t+H-1})$ over a control horizon $H$.
Because the environment is only partially observed, action prediction alone is insufficient.
The robot must anticipate how its viewpoint will change under motion and whether the resulting future state makes progress toward the goal.

\subsection{Visual Foresight and Action Selection}
\label{sec:visual_foresight_action_selection}
Under partial observability, successful navigation requires two coupled abilities: predicting how the view will change after moving, and choosing an action that brings the robot closer to the goal.
Direct policies learn this as a single action-prediction problem,
\begin{align}
    \pi_\theta(a_{t:t+H-1} \mid o_t, g).
\end{align}
This is efficient at test time, but the model is not explicitly trained to predict what it will see after executing the action.

NWMs make future prediction explicit by predicting future observations conditioned on candidate actions,
\begin{align}
    p_\theta(o_{t+H} \mid o_t, a_{t:t+H-1}).
\end{align}
However, they still require a separate action-selection procedure to choose which candidate action to execute.
This creates a gap between direct action prediction and explicit future prediction.

\subsection{From Navigation World Models to NavWAM}
\label{sec:from_nwm_to_navwam}
The NWM approach shows that future egocentric image prediction can support navigation by using CEM-based planning to sample candidate action sequences, predict their future observations, and select the best-scoring trajectory.
However, this pipeline still treats future prediction and action selection as separate steps: the model predicts possible futures, while an external procedure decides which future indicates goal progress and which action should be executed.
This separation increases test-time computation and makes closed-loop behavior depend on candidate sampling, scoring, and optimization.

\methodname{} addresses this separation by learning a joint world-action prediction,
\begin{align}
    p_\theta
    \left(
        a_{t:t+H-1},
        s_{t+H},
        o_{t+H-1:t+H},
        v_{t+H}
        \mid
        o_t, g
    \right).
\end{align}
Here, $s_{t+H}$ is an auxiliary future state, $o_{t+H-1:t+H}$ denotes the future egocentric observations predicted by the latent canvas, and $v_{t+H}$ is a goal-conditioned scalar value that estimates progress toward the goal.
At inference, \methodname{} predicts $\hat{a}_{t:t+H-1}$, $\hat{s}_{t+H}$, $\hat{o}_{t+H-1:t+H}$, and $\hat{v}_{t+H}$, and executes the predicted action chunk in a receding-horizon manner.
This formulation turns visual foresight into a closed-loop policy by learning future perception, goal progress, and executable motion as one coupled prediction problem.

\section{Method}
\label{sec:method}

\paragraph{Overview.}
\methodname{} parameterizes a closed-loop navigation policy using a pretrained video world model.
Rather than using the model only to predict future views for externally planned actions, \methodname{} represents the variables needed for navigation control as a shared latent canvas.
The current observation, image goal, robot state, action chunk, future egocentric observations, and goal-progress value are assigned to different latent frames in the same diffusion-transformer sequence.
This turns goal-conditioned navigation into a joint denoising problem over future perception, progress estimation, and executable motion.

\paragraph{Backbone.}
We instantiate \methodname{} with Cosmos Predict2~\citep{cosmospredict2}, a pretrained diffusion-transformer video world model.
Cosmos Predict2 provides the causal VAE~\cite{wan2025}, DiT backbone~\citep{Peebles_2023_ICCV}, and text-conditioning interface used to encode and denoise the latent sequence.
Following the latent-frame modeling principle of Cosmos Policy~\citep{cosmos_policy}, we encode non-visual robot variables as latent frames rather than attaching separate action or value heads.
This preserves the pretrained video-model interface while using a common latent representation for actions, states, values, and future images.
This design is navigation-specific: the value frame represents goal progress, and the action frame represents local-frame motion commands executed in a receding-horizon loop.
Unlike video policies developed primarily for manipulation, \methodname{} targets goal-conditioned navigation, where the model must couple viewpoint-changing egocentric prediction with goal-progress estimation and local-frame action generation.
Figure~\ref{fig:navwam_overview} summarizes the architecture.

\subsection{World-Action Latent Canvas}
\label{sec:latent_sequence}

\begin{wrapfigure}[13]{r}{0.58\textwidth}
\vspace{3pt}
\centering
\includegraphics[width=\linewidth]{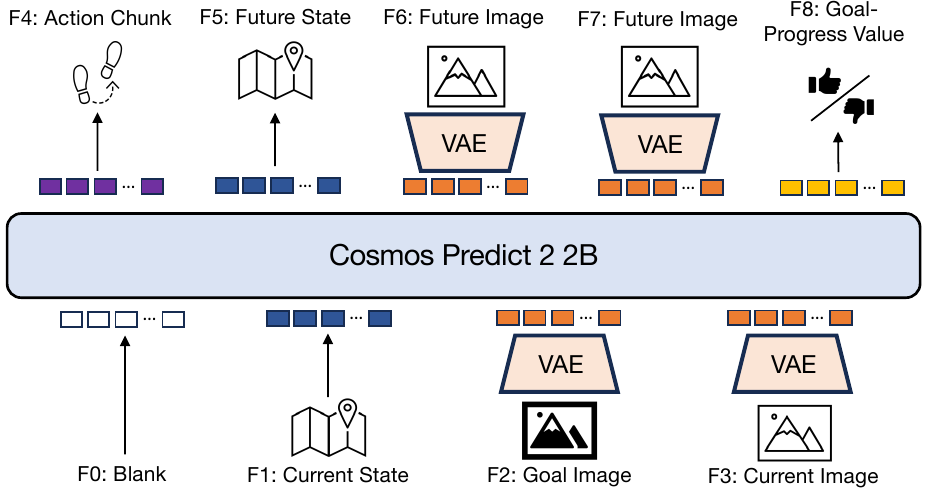}
\vspace{-12pt}
\caption{\methodname{} overview.}
\label{fig:navwam_overview}
\vspace{-0.25cm}
\end{wrapfigure}

\methodname{} represents navigation as denoising a fixed nine-frame latent canvas, as shown in Figure~\ref{fig:navwam_overview}.
The bottom four frames in the canvas are observed and condition the denoising process: a blank frame required by the causal VAE's temporal compression, the current robot state $s_t$, a goal frame containing the goal specification $g$, and the current egocentric observation $o_t$.
In our image-goal setting, $g$ is the target image captured at the goal location and is encoded as the goal-image frame.
The top five frames are generated as prediction targets: the executable action chunk $a_{t:t+H-1}$, the future state $s_{t+H}$, two future egocentric observations $o_{t+H-1}$ and $o_{t+H}$, and the goal-progress value $v_{t+H}$.
This canvas makes future-view prediction, progress estimation, and action generation part of the same denoising process, rather than separate modules.

Image frames are encoded as standard video latents through the causal VAE.
Non-image variables, including states, action chunks, and the scalar value, are normalized and broadcast over the latent spatial grid.
Their predictions are recovered by averaging the denoised entries of the corresponding frame.
This preserves the pretrained video-transformer interface while allowing one model to jointly predict visual and non-visual navigation variables.
Although the same canvas could in principle be extended to language- or object-specified goals through the goal frame or text-conditioning interface, we focus on image-goal navigation in this work.

\subsection{Training Objective}
\label{sec:training_objective}

Let $x_0$ denote the clean latent canvas and $x_\sigma$ its noisy version at noise level $\sigma$, obtained by adding Gaussian noise $\epsilon$ according to the diffusion schedule.
The diffusion transformer $F_\theta$ is trained to denoise the latent canvas with the objective
\begin{align}
    \mathcal{L}_{\mathrm{diff}}
    =
    \mathbb{E}_{\sigma,\epsilon}
    \left[
        w(\sigma)
        \left\|
            x_0 - F_\theta(x_\sigma,\sigma, c)
        \right\|_2^2
    \right],
\end{align}
where $w(\sigma)$ is the diffusion weighting term and $c$ denotes the conditioning information, including the observed-frame mask, conditioning embeddings, and observed latent frames.
The denoising loss is applied to the generated frames of the latent canvas.
We upweight the action frame relative to the other prediction frames so that the low-dimensional action signal is not dominated by the high-dimensional image reconstruction loss.

\methodname{} does not introduce separate action or value heads.
Instead, actions, states, and scalar values are decoded from their corresponding denoised latent frames.
Thus, action generation, future-state prediction, future-view prediction, and value estimation are trained as parts of the same world-action denoising objective, rather than as auxiliary losses added to an action-only policy.
For navigation, the value frame is trained to represent goal progress rather than a generic reward-to-go.
This encourages the model to estimate whether the predicted future state moves the robot closer to the specified goal.

\paragraph{Multi-mode Conditioning.}
The observed/generated frame pattern determines the training mode of each sample.
We use three modes: a policy mode that conditions on the current state, current observation, and goal, and predicts the action chunk, future state, future observations, and value; a world-model mode that additionally conditions on the action chunk and predicts the resulting future state, future observations, and value; and a value mode that conditions on the future frames and predicts the value.
In our main setting, samples are assigned to these modes with a $50/25/25$ split.
A single set of weights therefore learns action generation, action-conditioned future prediction, and value estimation.
All reported main results use only the policy mode at inference; the world-model mode supports optional best-of-$N$ sampling, and the value mode is trained as an auxiliary value-estimation mode.

\subsection{Inference}
\label{sec:inference}
At test time, \methodname{} runs in its policy mode.
Given the current egocentric observation $o_t$ and goal $g$, the model denoises the prediction frames and outputs a predicted action chunk $\hat{a}_{t:t+H-1}$.
The robot executes this chunk in a receding-horizon manner and then re-queries the model with the next observation.
Future observations and the goal-progress value are predicted alongside the action, but they are not required for execution.
Instead, they provide interpretable visual foresight and an estimate of whether the predicted future state moves toward the goal.

\methodname{} also supports optional value-guided best-of-$N$ sampling.
In this mode, the model draws $N$ candidate action chunks, evaluates their predicted futures and goal-progress values using the auxiliary world-model and value modes, and executes the chunk with the highest predicted value.
This optional mode is not used in the reported main results.
All reported main results use the default policy mode without CEM-style action search.
~\section{Experiments}
\label{sec:experiments}

\subsection{Experimental Setup}
\label{sec:exp_setup}
\vspace{-5pt}
\paragraph{Datasets.}
We evaluate image-goal navigation, where the agent receives a current egocentric observation and a target image, and must navigate to the location where the target image was captured.
We use \textsc{go stanford}~\citep{go_stanford_dataset} for NWM-style image-goal evaluation and construct additional image-goal episodes from \textsc{sit}~\cite{neurips_bae_2023} for held-out comparison with OmniVLA.
\textsc{go stanford} follows the protocol used by Navigation World Models~\citep{bar2024navigationworldmodels}, allowing direct comparison with NWM-style future-prediction-based planning.
We use \textsc{sit} for comparison with OmniVLA, since OmniVLA uses \textsc{go stanford} as part of its training data, making \textsc{sit} a cleaner held-out benchmark for direct-policy comparison.


\vspace{-5pt}
\paragraph{Baselines.}
We compare against three representative baselines.
NWM~\citep{bar2024navigationworldmodels} is our primary planning-based world-model baseline; it predicts future egocentric observations for sampled candidate actions and selects actions through external planning.
Cosmos Predict2~\citep{cosmospredict2} serves as a generic video-world-model planning baseline that predicts future observations and uses the same CEM-based action-selection protocol as NWM.
OmniVLA~\citep{omnivla} serves as a representative direct navigation policy that predicts actions without explicit future-view supervision.
For planning-based baselines, we follow the NWM-style CEM action-selection protocol, whereas all reported \methodname{} results use the default policy mode without best-of-$N$ sampling or CEM-style action search.
These baselines cover the main alternatives to \methodname{}: NWM-style planning, generic video-world-model planning, and direct action prediction.

\vspace{-5pt}
\paragraph{Metrics.}
For trajectory-level evaluation, we report Absolute Trajectory Error (ATE) and Relative Pose Error (RPE) at two evaluation horizons, 
$h \in \{4,8\}$, where lower is better.
When goal-reaching annotations are available, we report SR@$1.0$m.
For methods that predict future egocentric observations, we also report subject consistency~\citep{Liu_2025_ICCV}, the visual-feature similarity between predicted and ground-truth future observations.

\vspace{-5pt}
\paragraph{Real-world Evaluation Setup.}
%
We deploy \methodname{} and two baselines, NWM and OmniVLA, on a Diablo mobile robot equipped with an egocentric RGB camera.
All methods receive $224 \times 224$ RGB observations, output local-frame action commands, and are evaluated under the same onboard closed-loop control setting.
We evaluate 24 closed-loop image-goal navigation episodes per method across four indoor environments: Office, Storage, Meeting room, and Hallway.
Episodes terminate upon reaching within 1 meter of the goal or at a fixed timeout.

\subsection{Results}

\begin{figure}[t]
\centering
\begin{minipage}[t]{0.27\linewidth}
\vspace{0pt}
\centering
\centering
\footnotesize
\captionof{table}{\small Nav. performance}
\vspace{-5pt}
\label{tab:image_goal_nwm}
\begingroup
\setlength{\tabcolsep}{2.5pt}
\scriptsize
\begin{tabular}{@{}lcc@{}}
\toprule
Method & \textbf{ATE$\downarrow$} & \textbf{RPE$\downarrow$} \\
\midrule
Cosmos Predict2~\cite{cosmospredict2} & 0.455          & 0.109          \\
NWM~\cite{bar2024navigationworldmodels}             & 0.453          & 0.107          \\
NavWAM          & 0.324          & 0.099          \\
NavWAM w/ FT  & \textbf{0.192} & \textbf{0.070} \\
\bottomrule
\end{tabular}
\endgroup

\end{minipage}%
\hspace{0.015\linewidth}%
\begin{minipage}[t]{0.26\linewidth}
\vspace{0pt}
\centering
\input{fig/foresight_quality}
\end{minipage}%
\hspace{0.015\linewidth}%
\begin{minipage}[t]{0.41\linewidth}
\vspace{0pt}
\centering







\centering
\tiny
\captionof{table}{Head ablation of \methodname{}.}
\label{tab:ablation_targets}
\vspace{-5pt}
\begingroup
\setlength{\tabcolsep}{2.2pt}
\renewcommand{\arraystretch}{0.9}
\scriptsize
\begin{tabular}{@{}cccc l cc cc@{}}
\toprule
\multicolumn{4}{c}{\textbf{Sup. heads}}
& \multirow{2}{*}{\textbf{Infer.}}
& \multicolumn{2}{c}{\textbf{ATE $\downarrow$}}
& \multicolumn{2}{c}{\textbf{RPE $\downarrow$}} \\
\cmidrule(lr){1-4}
\cmidrule(lr){6-7}
\cmidrule(l){8-9}
Img. & Act. & St. & Val.
&
& $h{=}4$ & $h{=}8$
& $h{=}4$ & $h{=}8$ \\
\midrule

\checkmark &  &  & 
& planning
& 0.326 & 0.569
& 0.133 & 0.135 \\

\checkmark & \checkmark & \checkmark & 
& policy
& 0.107 & 0.287
& 0.054 & 0.098 \\

\checkmark & \checkmark & \checkmark & \checkmark
& policy
& \textbf{0.076} & \textbf{0.192}
& \textbf{0.037} & \textbf{0.070} \\

\bottomrule
\end{tabular}%
\endgroup

\end{minipage}%
\vspace{-10pt}
\end{figure}

\begin{figure}[t]
\begin{center}
\includegraphics[keepaspectratio, width=0.98\textwidth]{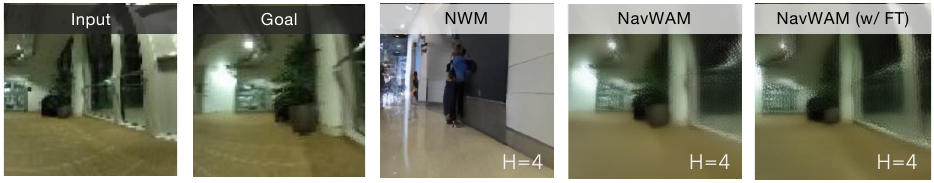}
\end{center}
\vspace{-0.3cm}
\caption{Qualitative future-view predictions on \textsc{go stanford}.}
\label{fig:qualitative_future_image}
\vspace{-10pt}
\end{figure}


\vspace{-5pt}
\paragraph{World Models as Policies.}
We first test whether \methodname{} can replace CEM-based future-prediction planning with direct action prediction.
Unlike Cosmos Predict2 and NWM, which sample candidate actions and select among predicted futures, \methodname{} directly predicts an action chunk in its default policy mode while also producing future observations and a goal-progress value.
Table~\ref{tab:image_goal_nwm} reports results on \textsc{go stanford}.
Without in-domain fine-tuning, \methodname{} reduces ATE compared with Cosmos Predict2 and NWM.
With a \textsc{go stanford} fine-tuning pass, \methodname{} achieves the best performance, reducing ATE to $0.192$ and RPE to $0.070$.
These results suggest that joint future, value, and action prediction can turn visual foresight into an effective navigation policy without CEM-style action search.

\vspace{-5pt}
\paragraph{Preserving Visual Foresight.}
We next ask whether converting a navigation world model into an action-producing policy compromises its ability to predict future egocentric observations.
Figure~\ref{fig:future_quality_bar} shows that \methodname{} preserves visual foresight while predicting actions directly.
Without task-specific fine-tuning, \methodname{} improves subject consistency over NWM, increasing it from $0.524$ to $0.668$.
After fine-tuning, subject consistency remains above NWM at $0.635$.
The qualitative examples in Figure~\ref{fig:qualitative_future_image} show a similar trend: \methodname{} predicts future views that remain closer to the goal scene, while NWM can drift to visually inconsistent futures.
These results suggest that visual foresight need not remain a separate CEM-based planning module; it can be integrated into an action-producing navigation policy while retaining consistent future-observation prediction.

\vspace{-5pt}
\paragraph{Learning Useful Futures for Control.}
We ablate the supervised prediction targets to test whether future-image prediction alone is sufficient for navigation.
Table~\ref{tab:ablation_targets} shows that image-only future prediction is not sufficient for navigation, even when combined with CEM using $N=120$ candidate actions.
Adding action and state supervision greatly improves trajectory accuracy, reducing ATE from $0.326/0.569$ to $0.107/0.287$ at horizons $h=4/8$.
Adding value supervision further improves performance, achieving the best ATE and RPE across both horizons.
These results suggest that useful visual foresight for navigation should be learned together with action and goal-progress prediction, rather than optimized as future-image prediction alone.

\begin{figure}[t]
\begin{center}
\includegraphics[keepaspectratio, width=\textwidth]{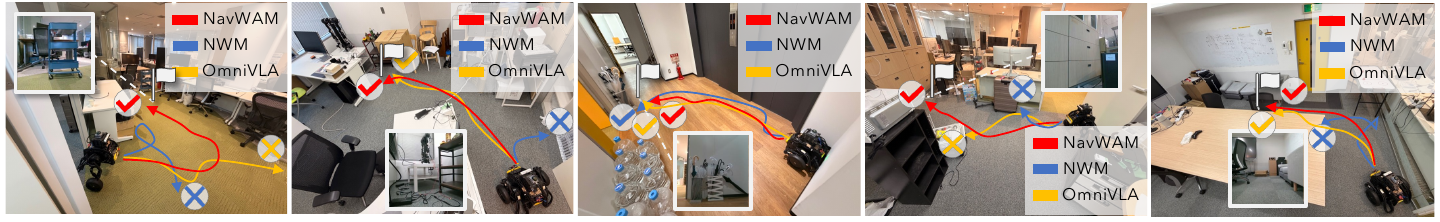}
\end{center}
\vspace{-0.3cm}
\caption{Real-world rollouts on the Diablo robot.}
\label{fig:realworld_rollout}
\vspace{-15pt}
\end{figure}

\begin{table}[t]
\centering
\begin{minipage}[t]{0.38\linewidth}
\vspace{0pt}
\centering
\caption{Direct-policy comparison.} 
\label{tab:vs_direct_policy}
\footnotesize
\setlength{\tabcolsep}{3pt}


{\renewcommand{\arraystretch}{0.85}
\begin{tabular}{l cc cc}
\toprule
& \multicolumn{2}{c}{ATE$\downarrow$} & \multicolumn{2}{c}{SR (\%) $\uparrow$} \\
\cmidrule(lr){2-3} \cmidrule(lr){4-5}
Method
& $h{=}4$ & $h{=}8$
& $h{=}4$ & $h{=}8$ \\
\midrule
OmniVLA & 0.086 & 0.162 & 45.4 & 12.1 \\
NavWAM  & \textbf{0.077} & \textbf{0.144} & \textbf{46.3} & \textbf{15.9} \\
\bottomrule
\end{tabular}
}

\end{minipage}%
\hfill
\begin{minipage}[t]{0.6\linewidth}
\vspace{0pt}
\centering
\caption{Real-world navigation performance.}
\label{tab:real_world_nav}
\footnotesize
\setlength{\tabcolsep}{3pt}

\begin{tabular}{lcccc|c}
\toprule
Method & Office & Storage & Meeting & Hallway & SR (\%) \\
\midrule
NWM~\citep{bar2024navigationworldmodels}         & 1/8 & 0/6 & 1/6 & 2/4 & 16.7\% \\
OmniVLA~\citep{omnivla} & 4/8 & 4/6 & 3/6 & \textbf{3/4} & 58.3\% \\
NavWAM (Ours)           & \textbf{6/8} & \textbf{6/6} & \textbf{4/6} & \textbf{3/4} & \textbf{79.2\%} \\
\bottomrule
\end{tabular}


\end{minipage}
\vspace{-10pt}
\end{table}

\vspace{-5pt}
\paragraph{Predictive Policies vs. Direct Policies.}
We compare \methodname{} with OmniVLA on \textsc{sit} to test whether explicit future prediction degrades direct action prediction.
Table~\ref{tab:vs_direct_policy} shows that \methodname{} remains competitive with OmniVLA on \textsc{sit}, with lower ATE and slightly higher observed success rates at both horizons.
These results are obtained with a 2B-parameter video backbone, whereas OmniVLA is built on a 7B-parameter OpenVLA backbone.
Thus, \methodname{} provides action accuracy comparable to a larger direct navigation policy while additionally producing future observations and goal-progress values.

\begin{figure}[h]
\begin{center}
\includegraphics[keepaspectratio, width=\textwidth]{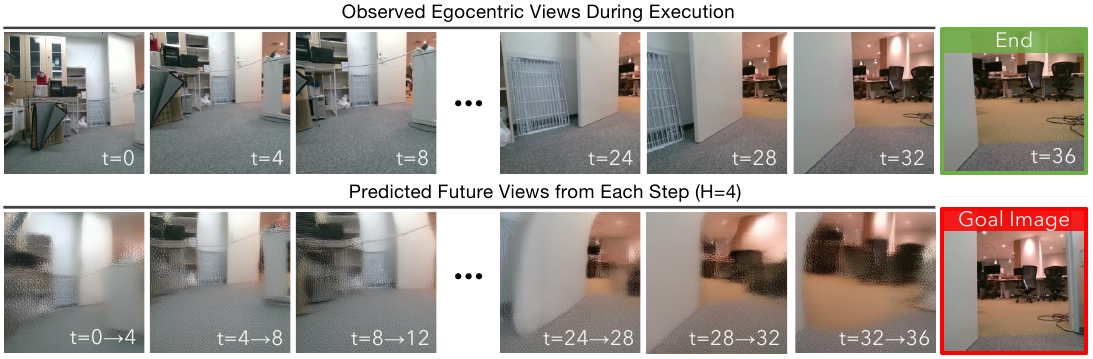}
\end{center}
\vspace{-10pt}
\caption{Real-world future-view prediction by \methodname{} during closed-loop execution.}
\label{fig:realworld_ftr}
\vspace{-10pt}
\end{figure}

\vspace{-5pt}
\paragraph{Closed-loop Real-Robot Deployment.}
We evaluate real-world deployment on a Diablo robot across 24 closed-loop image-goal episodes in four indoor environments.
As shown in Table~\ref{tab:real_world_nav}, \methodname{} obtains the highest observed success rate, reaching the goal in $19/24$ episodes ($79.2\%$; 95\% Wilson CI: $[59.5, 90.8]$), compared with $14/24$ for OmniVLA ($58.3\%$; 95\% CI: $[38.8, 75.5]$) and $4/24$ for NWM ($16.7\%$; 95\% CI: $[6.7, 35.9]$).
Figure~\ref{fig:realworld_rollout} shows representative rollouts where \methodname{} reaches the goal region more consistently, while NWM often drifts and OmniVLA sometimes stops short or follows less direct paths.
Figure~\ref{fig:realworld_ftr} further illustrates action-consistent visual foresight during real-world execution: future views predicted at each step qualitatively resemble the subsequent egocentric observations after the robot executes the predicted action chunks.
Although the number of trials is limited, these results suggest that jointly learning future prediction, goal-progress estimation, and action generation can transfer to closed-loop real-robot image-goal navigation while preserving interpretable visual foresight.

~

\vspace{-5pt}
\section{Conclusion} \label{sec:conclusion}
We presented \methodname{}, a navigation world action model that turns visual foresight into a closed-loop policy for goal-conditioned visual navigation.
By representing future observations, goal-progress values, and executable action chunks in a shared latent sequence, \methodname{} learns future perception, progress estimation, and action generation as one coupled prediction problem.
In our evaluations, this joint formulation improves navigation performance over planning-based world-model baselines without CEM-style action search in the default policy mode, remains competitive with a larger direct navigation policy, and transfers to closed-loop real-robot image-goal navigation.
These results suggest that future prediction is most useful for robot navigation when it is learned together with the action and value targets that determine closed-loop behavior, rather than treated as a separate planning module.

\vspace{-5pt}
\paragraph{Limitations.}
Our evaluation focuses on image-goal navigation in indoor environments, and broader evaluation on language- or object-specified goals remains future work.
Our real-world study is limited in scale, covering 24 closed-loop episodes across four environments.
All reported main results use the default policy mode, and studying when optional value-guided best-of-$N$ sampling improves robustness is left for future work.

\if[]
\acknowledgments{If a paper is accepted, the final camera-ready version will (and probably should) include acknowledgments. All acknowledgments go at the end of the paper, including thanks to reviewers who gave useful comments, to colleagues who contributed to the ideas, and to funding agencies and corporate sponsors that provided financial support.}
\fi

\bibliography{main}

\clearpage
~\appendix
{\Large\textbf{Appendix}\par}
\renewcommand{\thefigure}{S\arabic{figure}}
\renewcommand{\thetable}{S\arabic{table}}
\setcounter{figure}{0}
\setcounter{table}{0}
~\section{Method Details}
\label{sup:method}

The main paper specifies \methodname{} at the level of a shared latent canvas, a diffusion-transformer policy, and a simulation-pretrain plus real-robot adaptation recipe. This section fixes the concrete definitions, training curriculum, and hyperparameters needed to reproduce that recipe end to end, without consulting the source code.

\subsection{Latent Canvas Frame Layout}
\label{sup:frame_layout}

\methodname{} represents goal-conditioned navigation as denoising a fixed nine-frame latent canvas. Each frame is encoded into the shared video-latent space, broadcast over the latent spatial grid if its variable is non-image, and either treated as observed (conditioning) or generated (prediction target). Table~\ref{tab:sup_frame_layout} summarizes the layout used in all reported experiments. The causal VAE backbone applies a $4{:}1$ temporal compression, so the $9$-frame latent canvas corresponds to $1$ pad frame plus $8 \times 4 = 32$ raw input frames, i.e., a VAE temporal chunk of $33$ raw frames.

\begin{table}[h]
\centering
\caption{\textbf{Latent canvas frame layout of \methodname{}.} The same nine-frame canvas is shared by all training phases and by inference; Frame~2 carries the goal image or the previous-step FPV.}
\label{tab:sup_frame_layout}
\setlength{\tabcolsep}{4pt}
\renewcommand{\arraystretch}{0.95}
\begin{tabular}{c l l c c}
\toprule
Frame & Content & Dim. & Observed & Predicted \\
\midrule
0 & blank (causal-VAE temporal pad)                      & ---                   & yes & --- \\
1 & current state $s_t = [x, y, \psi]$                   & 3                     & yes & --- \\
2 & goal image $g$ \emph{(IG)} or $o_{t-1}$ \emph{(WM)} & $3{\times}224{\times}224$ & yes & --- \\
3 & current observation $o_t$                            & $3{\times}224{\times}224$ & yes & --- \\
\midrule
4 & action chunk $a_{t:t+H-1}$                           & $3H$                  & --- & yes \\
5 & future state $s_{t+H}$                               & 3                     & --- & yes \\
6 & future observation $o_{t+H-1}$                       & $3{\times}224{\times}224$ & --- & yes \\
7 & future observation $o_{t+H}$                         & $3{\times}224{\times}224$ & --- & yes \\
8 & goal-progress value $v_{t+H} \in [0,1]$              & 1                     & --- & yes \\
\bottomrule
\end{tabular}
\end{table}

We use two variants of the canvas that differ only in the content of Frame~2:

\begin{itemize}[leftmargin=1.2em,itemsep=2pt,topsep=2pt]
    \item \textbf{IG variant (image-goal navigation).} Frame~2 holds the goal image $g$, sampled from a future step on the same trajectory at training time and provided by the operator at deployment.
    \item \textbf{WM variant (world-model variant, used for foresight quality and language-goal navigation).} Frame~2 holds the previous-step egocentric observation $o_{t-1}$; the goal specification is supplied through the text-conditioning interface of the video backbone.
\end{itemize}

Non-image variables (state, action chunk, scalar value) are normalized and broadcast over the latent spatial grid; their predictions are recovered by spatially averaging the denoised entries of the corresponding frame.

\subsection{Goal-Progress Value \texorpdfstring{$v_{t+H}$}{v(t+H)}}
\label{sup:value_def}

The value frame encodes a bounded goal-progress estimate. For the joint real-robot fine-tuning datasets (\textsc{recon}, \textsc{sacson}, \textsc{scand}) and the in-domain \textsc{go stanford} fine-tune, we use
\begin{align}
    v_{t+H}
    \;=\;
    \mathrm{clip}\!\left(
        1 - \frac{\bigl\| p_\text{end} - p_t \bigr\|_{2}}{d_{\max}},
        \;0,\;1
    \right),
    \label{eq:sup_value}
\end{align}
where $p_t$ is the current 2D position, $p_\text{end}$ is the final 2D position of the recorded trajectory, and $\|\cdot\|_2$ is Euclidean distance. Labels are read directly from the trajectory metadata bundled with each NWM-style dataset. The cap $d_{\max}$ is set to the upper percentile of trajectory lengths in our training data, ensuring that the value remains a bounded scalar in $[0,1]$.

For the \textsc{hm3d} simulation pretrain we use a geodesic version of Eq.~\eqref{eq:sup_value}, computed from the Habitat shortest-path API on the simulator scene mesh. The bounded $[0,1]$ form is unchanged.

\subsection{Robot State \texorpdfstring{$s_t$}{s(t)}}
\label{sup:state_def}

The state frame encodes a 3D vector
\begin{align}
    s_t \;=\; \bigl[\, x_t / 100,\;\; y_t / 100,\;\; \psi_t / \pi \,\bigr] \in \mathbb{R}^3,
\end{align}
where $(x_t, y_t)$ are 2D positions in the trajectory's recorded frame (meters) and $\psi_t$ is the yaw (radians). The coarse normalization keeps the state numerically in a similar range to the other normalized frames; the model treats $s_t$ as a 3-entry vector broadcast over the latent spatial grid.

\subsection{Action Chunk \texorpdfstring{$a_{t:t+H-1}$}{a(t:t+H-1)}}
\label{sup:action_def}

Each action chunk contains $H$ local-frame waypoint increments
\begin{align}
    a_i \;=\; \bigl(\Delta x_i,\; \Delta y_i,\; \Delta \psi_i\bigr) \in \mathbb{R}^3,
    \qquad i = t,\ldots,t+H-1,
\end{align}
expressed in the robot frame at time $t$. Translational components are normalized by a per-dataset waypoint spacing and then linearly rescaled to $[-1, +1]$ together with $\Delta \psi$ (concrete spacings are given in Supp.~\ref{sup:numerical_hp}). The action frame thus has dimension $3H$ which is broadcast over the latent spatial grid and decoded by spatial averaging.

We use chunk size $H{=}4$ throughout the main paper. The \textsc{hm3d} simulation pretrain uses a longer chunk $H{=}16$, which is reset to $H{=}4$ at the start of the joint fine-tune (Supp.~\ref{sup:curriculum}). Online, the chunk is executed in a receding-horizon manner: the robot consumes the chunk at the dataset-native rate and re-queries the model when the chunk is exhausted.

\subsection{Training Objective and Conditioning Modes}
\label{sup:training_objective}

Following the main paper, the diffusion transformer $F_\theta$ denoises the latent canvas under the loss
\begin{align}
    \mathcal{L}_{\mathrm{diff}}
    \;=\;
    \mathbb{E}_{\sigma,\epsilon}
    \!\left[
        w(\sigma)\,
        \bigl\|\, x_0 - F_\theta(x_\sigma,\sigma,c) \,\bigr\|_{2}^{\!2}
    \right],
\end{align}
applied only to the generated frames of the canvas. The action frame is upweighted relative to the future-image frames so that the low-dimensional action signal is not dominated by the per-pixel reconstruction loss; the concrete multiplier is given in Supp.~\ref{sup:numerical_hp}.

\paragraph{Three Conditioning Modes.}
At each training step we sample one of three observed/generated patterns:
\begin{itemize}[leftmargin=1.2em,itemsep=1pt,topsep=2pt]
    \item \emph{Policy mode} ($50\%$): condition on Frames 0--3, predict Frames 4--8 (action, future state, future images, value).
    \item \emph{World-model mode} ($25\%$): also condition on Frame 4 (action), predict Frames 5--8.
    \item \emph{Value mode} ($25\%$): condition on Frames 0--7, predict Frame 8 (scalar value).
\end{itemize}
The mixture is sampled \emph{per training sample}, so a single set of weights learns action generation, action-conditioned future prediction, and value estimation. All reported main results use the policy mode at inference; the world-model and value modes are auxiliary modes used only during training.
~\section{Implementation Details}
\label{sup:implementation_details}

This section describes the training hyperparameters, training curriculum, dataset configurations, and real-robot platform used to reproduce \methodname{} end to end without consulting the source code.

\subsection{Training Hyperparameters}
\label{sup:numerical_hp}

\paragraph{Action Loss Weight.}
The diffusion objective of Supp.~\ref{sup:training_objective} upweights the action frame relative to the future-image frames by a multiplier $\lambda{=}5$.

\paragraph{Action Normalization.}
Translational components of each action chunk are first divided by a per-dataset waypoint spacing: $0.25$\,m for \textsc{recon}, $0.255$\,m for \textsc{sacson}, $0.38$\,m for \textsc{scand}, and $0.12$\,m for \textsc{go stanford}. The resulting $(\Delta x, \Delta y, \Delta \psi)$ triple (with $\Delta \psi$ in radians) is linearly rescaled to $[-1, +1]$ before being broadcast over the latent spatial grid.

\paragraph{Noise Schedule and Scaling.}
Denoising uses the rectified-flow scaling of the Cosmos Predict2 backbone (\texttt{scaling=rectified\_flow}, with data-noise scale $\sigma_{\text{data}}{=}1.0$ and conditioning-frame noise $\sigma_{\text{cond}}{=}0$). Training noise is drawn from a hybrid distribution with $\sigma_{\max}{=}200$, $\sigma_{\min}{=}10^{-2}$, a log-normal core $(p_{\text{mean}}{=}1.39, p_{\text{std}}{=}1.2)$, and a uniform tail on $[1,85]$. Inference uses a narrower range, $\sigma_{\max}{=}80$, $\sigma_{\min}{=}4$.

\subsection{Training Curriculum}
\label{sup:curriculum}

We train \methodname{} in three phases. Each phase initializes its weights from the previous phase and writes a single checkpoint that we use throughout the main paper.

\paragraph{Phase 1: \textsc{hm3d} Simulator Pretrain.}
We train on the success-only split of \textsc{hm3d} simulator trajectories generated by our navigation policy, with rollout demonstrations sampled with probability $0.5$ alongside expert trajectories. The \textsc{hm3d} phase uses a long chunk ($H{=}16$) to expose the model to long-horizon visual foresight from the simulator.

\paragraph{Phase 2: Joint Fine-Tune on Real-Robot Data.}
The Phase~1 checkpoint is fine-tuned jointly on \textsc{recon}, \textsc{sacson}, and \textsc{scand}, using NWM-style continuous actions with chunk size $H{=}4$. Because the action frame is repeat-filled within Frame~4 of the latent canvas, changing $H$ between phases does not alter the canvas dimensions or require re-initialization of the model weights. Both the WM and IG variants of \methodname{} branch off this stage: they share Phase~1 and differ only in the content of Frame~2 during Phase~2. The resulting WM and IG variants are what the main paper calls \emph{\methodname{} (zero-shot)} on \textsc{go stanford}, since \textsc{go stanford} is absent from Phase~2 training.

\paragraph{Phase 3: Optional Per-Dataset Fine-Tune.}
For results labelled \emph{\methodname{} w/ FT} on \textsc{go stanford}, we additionally fine-tune the Phase~2 IG checkpoint on the \textsc{go stanford} training trajectories with the same schedule.

\subsection{Datasets}
\label{sup:datasets}

\paragraph{Image-Goal Navigation.}
\textsc{go stanford}~\citep{go_stanford_dataset} provides the indoor image-goal evaluation. We hold out $30$ episodes as the test split and use the remaining trajectories for Phase~3 fine-tuning. The 30-episode test split is the largest size on which the NWM-style CEM planning protocol~\citep{bar2024navigationworldmodels} with $N{=}120$ candidates can be reproduced within our compute budget. The held-out image-goal evaluation on \textsc{sit}~\cite{neurips_bae_2023} follows the protocol of the main paper, using the $14$-segment official test split with $100$ episodes per segment for a total of $1{,}400$ episodes; per-segment counts are listed in Table~\ref{tab:sup_datasets}.

\paragraph{Training Data.}
Phase~1 uses the success-only split of \textsc{hm3d} simulator trajectories generated by our navigation policy. Phase~2 uses the public \textsc{recon}, \textsc{sacson}, and \textsc{scand} splits used by NWM. Phase~3 uses the \textsc{go stanford} training split; the same trajectories used for evaluation are never seen during fine-tune.

\begin{table}[ht]
\centering
\caption{\textbf{Dataset summary.} Trajectory counts (training) and episode counts (offline and closed-loop evaluation) for the data used in this paper. The \textsc{go stanford} offline evaluation uses the fixed 30-episode subset adopted in the main paper.}
\label{tab:sup_datasets}
\setlength{\tabcolsep}{4pt}
\renewcommand{\arraystretch}{0.95}
\begin{tabular}{l l l l}
\toprule
Role & Dataset & Split & Count \\
\midrule
Phase 1 train     & \textsc{hm3d}~\cite{hm3d}                                  & added + collected           & $802$ scenes / $185{,}000$ trajectories \\
Phase 2 train     & \textsc{recon}~\cite{recon_dataset}                        & train                       & $11{,}835$ trajectories \\
Phase 2 train     & \textsc{sacson}~\cite{sacson_dataset}                       & train                       & $2{,}000$ trajectories \\
Phase 2 train     & \textsc{scand}~\cite{scand_dataset}                        & train                       & $372$ trajectories \\
Fine Tuning & \textsc{go stanford}~\cite{go_stanford_dataset}                 & train                       & $3{,}544$ trajectories \\
\midrule
Offline eval   & \textsc{go stanford}~\cite{go_stanford_dataset}            & test & 30 episodes \\
Offline eval   & \textsc{sit}~\cite{neurips_bae_2023}                          & test & $14$ segments / $1{,}400$ episodes \\
Closed-loop eval  & - & real-world rollouts  & 24 episodes \\
\bottomrule
\end{tabular}
\end{table}

\subsection{Hyperparameter Summary}
\label{sup:hyperparams}

Table~\ref{tab:sup_hyperparams} consolidates the values used in all phases of the main paper. The optimizer, learning rate, batch size, mode mixture, action upweight, and noise schedule are constant across phases; only the dataset, chunk size, and step count change.

\begin{table}[h]
\centering
\scriptsize
\caption{\textbf{Hyperparameters used in all phases of \methodname{}.} Values are identical across phases unless explicitly noted. Phase-specific differences (dataset, chunk size, step count, warmup) are listed in Supp.~\ref{sup:curriculum}.}
\label{tab:sup_hyperparams}
\setlength{\tabcolsep}{4pt}
\renewcommand{\arraystretch}{0.95}
\begin{tabular}{l l}
\toprule
Item & Value \\
\midrule
Backbone                            & Cosmos-Predict2 2B Video2World, $480$p, $16$\,fps \\
Resolution                          & $224 \times 224$ \\
Latent canvas                       & $9$ frames (Table~\ref{tab:sup_frame_layout}) \\
VAE temporal chunk duration         & $33$ raw frames ($1$ pad $+$ $8$ latent frames $\times$ $4{:}1$ causal-VAE compression) \\
\midrule
Scaling                             & Rectified flow (data-noise $\sigma_{\text{data}}{=}1.0$, conditioning $\sigma_{\text{cond}}{=}0$) \\
Training noise SDE                  & Hybrid EDM, $\sigma_{\max}{=}200$, $\sigma_{\min}{=}10^{-2}$ \\
Training noise log-normal core      & $p_{\text{mean}}{=}1.39$, $p_{\text{std}}{=}1.2$, uniform tail $[1, 85]$ \\
Inference noise SDE                 & $\sigma_{\max}{=}80$, $\sigma_{\min}{=}4$ \\
Conditioning strategy               & frame-replace, conditional frames denoised with GT (\texttt{denoise\_replace\_gt\_frames=true}) \\
Future-frame loss mask              & off (\texttt{mask\_loss\_for\_action\_future\_state\_prediction=false}) \\
\midrule
Mode mixture (policy/WM/value)      & $50\% / 25\% / 25\%$ per sample \\
Action loss multiplier $\lambda$    & $5$ \\
\midrule
Optimizer                           & AdamW \\
Learning rate                       & $10^{-4}$ \\
Per-GPU batch size                  & $8$ \\
GPUs per training run               & $4$ (RTX PRO 6000) \\
Effective batch                     & $32$ \\
LR scheduler                        & cosine with linear warmup, $f_\text{start}{=}10^{-6}$, $f_\text{max}{=}1$, $f_\text{min}{=}0.3$ \\
EMA                                 & disabled \\
Mixed precision                     & bfloat16 \\
\midrule
\textsc{hm3d} rollout sampling probability   & $0.5$ (Phase 1 only) \\
Waypoint spacing (m)                & \textsc{recon} $0.25$, \textsc{sacson} $0.255$, \textsc{scand} $0.38$, \textsc{go stanford} $0.12$ \\
\bottomrule
\end{tabular}
\end{table}

\subsection{Robot Platform}
\label{sup:robot_platform}

\begin{figure}[htbp]
\centering
\includegraphics[keepaspectratio, scale=0.40]{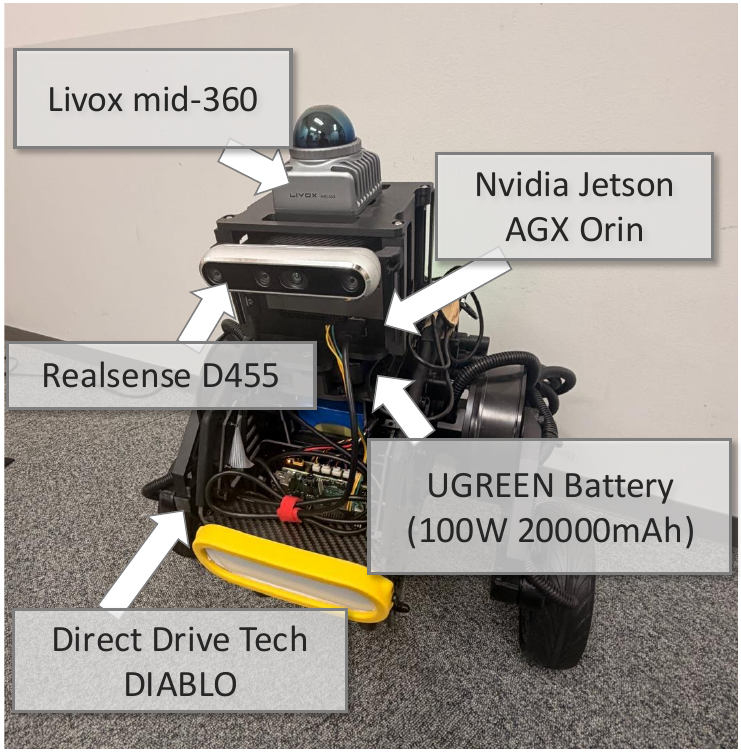}
\caption{\textbf{Robot platform.} Direct Drive Tech \emph{Diablo} with a 3D-printed frame carrying the onboard sensors and compute.}
\label{fig:sup_diablo}
\end{figure}

Figure~\ref{fig:sup_diablo} shows the robot platform used for our real-world experiments. The base is a Direct Drive Tech \emph{Diablo} fitted with a 3D-printed frame that holds an Intel RealSense D455 RGB-D camera, a Livox Mid-360 LiDAR, an NVIDIA Jetson AGX Orin, and a dedicated battery for the Orin and the LiDAR. The platform is controlled through ROS~2 via standard \texttt{cmd\_vel} commands, which we use to execute the local-frame action chunks predicted by \methodname{} during deployment.

~\section{Additional Results}

\subsection{Inference Efficiency}
\label{sup:efficiency}

We quantify the main-paper claim that \methodname{} replaces CEM-style trajectory optimization with a single denoising chain by jointly tabulating trajectory accuracy and inference cost against world-model planners on \textsc{go stanford} (Table~\ref{tab:sup_inference_efficiency}).

\paragraph{Compared Policies.}
We compare three policies. NWM~\citep{bar2024navigationworldmodels} uses a $1$B-parameter CDiT-XL backbone and selects actions via Cross-Entropy Method (CEM) search: at each step it samples $N$ candidate action sequences, predicts their future observations, scores them against the goal image, and executes the first action of the best-scoring trajectory. To isolate the contribution of the CEM search from that of the backbone, we add a Cosmos Predict2 + CEM baseline that wraps the same CEM scheme around the $2$B Cosmos Predict2~\citep{cosmospredict2} backbone used by \methodname{}. \methodname{} itself uses the same $2$B Cosmos Predict2 backbone but replaces the CEM search with a single denoising chain over the nine-frame latent canvas. For NWM we sweep the CEM budget $N \in \{8, 16, 32, 64, 120, 240\}$; $N{=}120$ matches the headline number in the main paper.

\paragraph{Measurement Setup.}
All numbers are measured on a single NVIDIA Blackwell-class GPU (RTX PRO 6000, $96$\,GB) in bfloat16 with $224\times224$ RGB inputs, averaged over $100$ closed-loop steps after a $20$-step warm-up. Peak GPU memory is the maximum of \texttt{torch.cuda.max\_memory\_allocated} over the same window. Trajectory metrics are reported on the same 30-episode subset of \textsc{go stanford} used in the main paper. When the action horizon $H$ differs across methods we normalize cost \emph{per executed action}: for the CEM baselines this covers both candidate rollouts and goal-image scoring; for \methodname{} it covers the full denoising chain. For NWM and Cosmos Predict2 + CEM, the $N$ candidates are batched into a single forward pass. FLOPs are given in TF ($1\,\text{TF}{=}10^{12}$\,FLOPs).

\paragraph{Cost--Accuracy Comparison.}
Three observations stand out in Table~\ref{tab:sup_inference_efficiency}. \emph{(i)~CEM accuracy saturates well below the paper-faithful budget}: NWM's trajectory metrics are essentially flat from $N{=}8$ onwards, and doubling $N$ from $120$ to $240$ slightly degrades accuracy rather than closing the gap to single-pass \methodname{}. \emph{(ii)~Single-pass \methodname{} dominates the entire CEM curve}: zero-shot \methodname{} matches or outperforms NWM at every $N$, and in-domain fine-tuning yields an additional $\sim\!1.7\times$ improvement in ATE. \emph{(iii)~The cost separation is structural}: CEM cost scales linearly with the candidate count $N$, whereas \methodname{} uses a single denoising chain independent of any candidate budget. Consequently, the same-backbone Cosmos Predict2 + CEM baseline requires orders of magnitude more compute per executed action than \methodname{}, with peak GPU memory following the same trend ($4{-}10\times$ larger for the CEM baselines). In wall-clock terms, \methodname{} runs at roughly $5$\,Hz on the same hardware, comfortably within a real-time control budget, whereas the CEM baselines run at sub-Hz rates under the paper-faithful $N{=}120$ setting. Table~\ref{tab:sup_inference_efficiency} reports the full numbers. Together, these trends support the main-paper claim that joint world--action denoising removes the need for an external CEM search, both in trajectory accuracy and in the cost of running the policy on hardware.

\begin{table}[t]
\centering
\footnotesize
\caption{\textbf{Inference efficiency and accuracy against world-model planners on \textsc{go stanford}.} Trajectory accuracy at $H{=}8$ on the same 30-episode subset as the main paper. \methodname{} replaces the CEM search loop with a single denoising chain; CEM baselines sweep $N$ candidate trajectories per executed action, with $N{=}120$ matching the NWM main-paper protocol. Best per column in \textbf{bold}.}
\label{tab:sup_inference_efficiency}
\setlength{\tabcolsep}{2pt}
\renewcommand{\arraystretch}{0.95}
\begin{tabular}{l l rr r r r}
\toprule
Method & Inference & ATE $\downarrow$ & RPE $\downarrow$ & FLOPs/act [TF] $\downarrow$ & Latency [ms] $\downarrow$ & Peak GPU [GB] $\downarrow$ \\
\midrule
\multirow{6}{*}{NWM}
& CEM, $N{=}8$            & 0.464 & 0.109 & $1{,}005$  & $26{,}168$  & 17.45 \\
& CEM, $N{=}16$           & 0.454 & 0.108 & $1{,}970$  & $37{,}287$  & 26.81 \\
& CEM, $N{=}32$           & 0.460 & 0.109 & $3{,}901$  & $69{,}737$  & 44.32 \\
& CEM, $N{=}64$           & 0.456 & 0.108 & $7{,}761$  & $127{,}752$ & 33.19 \\
& CEM, $N{=}120$ & 0.452 & 0.107 & $14{,}521$ & $233{,}831$ & 51.65 \\
& CEM, $N{=}240$          & 0.470 & 0.110 & $28{,}993$ & $469{,}320$ & 52.53 \\
\midrule
Cosmos Predict2
& CEM, $N{=}120$          & 0.455 & 0.109 & $18{,}114$ & $887{,}606$ & 20.04 \\
\midrule
\multirow{2}{*}{\textbf{\methodname{} (2B)}}
& zero-shot  & 0.324 & 0.099 & $4.45$     & \textbf{205.7} & \textbf{4.82} \\
& w/ FT   & \textbf{0.192} & \textbf{0.070} & \textbf{4.45} & \textbf{205.7} & \textbf{4.82} \\
\bottomrule
\end{tabular}
\end{table}

\subsection{Component Ablation}
\label{sup:component_ablation}

We complement the main-paper finding that image-only future prediction with CEM is not a competitive policy with the converse question: starting from a strong action--state--value policy, does \emph{adding} future-view supervision improve closed-loop control? Together with the main-paper result, the two ablations bracket the joint formulation from both sides.

\paragraph{Setup.}
To avoid confounding capacity and compute with the prediction target, both rows of Table~\ref{tab:sup_component_ablation} share the same nine-frame latent canvas, the same Cosmos Predict2 2B backbone, the same Phase~1+2+3 curriculum, the same datasets, the same optimizer, and the same number of training steps. The variants differ only in which generated frames receive a denoising loss: in Row~1 (without future-view) Frames~6/7 are still produced by the model but their loss term is zeroed out, so any performance gap is attributable to the supervision signal rather than to model capacity. All numbers are on the same 30-episode \textsc{go stanford} image-goal subset as the main paper, at evaluation horizons $h\in\{4,8\}$ and with a single forward pass.

\begin{table}[t]
\centering
\caption{\textbf{Effect of future-view supervision on \textsc{go stanford} image-goal ($n{=}30$).} The two rows share the same backbone, canvas, curriculum, datasets, and step budget; they differ only in whether Frames~6/7 receive a denoising loss. Row~2 (bold) is reproduced from Table~2 of the main paper. Best per column in \textbf{bold}.}
\label{tab:sup_component_ablation}
\setlength{\tabcolsep}{3pt}
\renewcommand{\arraystretch}{0.95}
\begin{tabular}{l cccc cc cc}
\toprule
& \multicolumn{4}{c}{Sup.~heads} & \multicolumn{2}{c}{ATE~$\downarrow$} & \multicolumn{2}{c}{RPE~$\downarrow$} \\
\cmidrule(lr){2-5} \cmidrule(lr){6-7} \cmidrule(lr){8-9}
Row & Img.\,fut. & Act. & St. & Val. & $h{=}4$ & $h{=}8$ & $h{=}4$ & $h{=}8$ \\
\midrule
NavWAM (wo/ Future Image)            & --- & \checkmark & \checkmark & \checkmark & 0.090 & 0.262 & 0.045 & 0.103 \\
\textbf{NavWAM}          & \checkmark & \checkmark & \checkmark & \checkmark & \textbf{0.076} & \textbf{0.192} & \textbf{0.037} & \textbf{0.070} \\
\bottomrule
\end{tabular}
\end{table}

\paragraph{Results.}
Table~\ref{tab:sup_component_ablation} shows that turning the future-view denoising loss back on (Row~2) improves trajectory accuracy at both evaluation horizons. At the long horizon $h{=}8$, ATE drops from $0.262$ to $0.192$ ($-27\%$) and RPE from $0.103$ to $0.070$ ($-32\%$); at the short horizon $h{=}4$ the improvement is in the same direction ($0.090\!\to\!0.076$ ATE, $0.045\!\to\!0.037$ RPE). The improvement is therefore not a single-horizon artifact: future-view supervision helps the policy at both short and long lookaheads. Combined with the main-paper finding that image-only future prediction with CEM is not a competitive policy, the result establishes that future-view prediction is neither sufficient on its own nor redundant once paired with action and value supervision---it is the joint formulation, rather than any single prediction target, that makes visual foresight usable for control.

\paragraph{Why Future-View Supervision Helps.}
Two mechanisms account for the gain. \emph{(i)~Dense auxiliary supervision improves the backbone representation:} Sparse action/state/value targets provide little direct supervision over observation space, whereas the future-view loss restores a dense reconstruction signal aligned with the Cosmos Predict2 video-generation objective. This helps keep the backbone close to its pretrained representation and improves the downstream action/value heads. \emph{(ii)~Future-view prediction internalizes the goal-conditioned transition under partial observability:} Since the goal is given as a single distant frame, the agent must choose actions whose resulting observations are not yet visible. Predicting Frames~6/7 forces the model to maintain an internal estimate of what should be seen a few steps ahead, providing the action head with a goal-consistent visual anchor. The ATE reduction is larger at the longer horizon ($-27\%$ at $h{=}8$ vs.\ $-16\%$ at $h{=}4$), which is precisely the regime where explicit future prediction should be most beneficial.

\subsection{Real-Robot Failure Modes}
\label{sup:rw_failures}

\paragraph{Protocol.}
For each of the $24$ image-goal episodes in the closed-loop deployment, the operator selected a start position and a goal image, and all three methods were then run from the same start with the same goal image, in randomized order within the session. For each (episode, method) pair we record a binary outcome ($1$ = reached within $1$\,m of the operator-marked goal, $0$ = otherwise), and each failed pair is labeled with one of two failure modes:
\begin{itemize}[leftmargin=1.2em,itemsep=1pt,topsep=2pt]
    \item \textbf{Drift}: the robot leaves the goal-image-implied scene and stops or wanders elsewhere.
    \item \textbf{Collision}: the robot contacts an obstacle or is stopped by the operator for safety.
\end{itemize}
Labels are assigned by inspection of the on-board video together with the recorded action stream; a single label is assigned per failed episode according to the dominant mode. Table~\ref{tab:sup_failure_modes} aggregates these labels by environment and method.

\begin{table}[t]
\centering
\footnotesize
\caption{\textbf{Failure-mode breakdown of the 24 Diablo closed-loop episodes.} Each failure is labeled with one of two modes (drift, collision). The \#Episodes column gives the per-environment episode budget. The bottom block reports the aggregate per-method counts.}
\label{tab:sup_failure_modes}
\setlength{\tabcolsep}{4pt}
\renewcommand{\arraystretch}{0.95}
\begin{tabular}{l c l | c c | c}
\toprule
Environment & \#Episodes & Method & Drift & Collision & Success \\
\midrule
\multirow{3}{*}{Office} & \multirow{3}{*}{8}
& NWM      & 3 & 4 & 1/8 \\
& & OmniVLA  & 2 & \textbf{2} & 4/8 \\
& & \methodname{} & \textbf{0} & \textbf{2} & \textbf{6/8} \\
\midrule
\multirow{3}{*}{Storage} & \multirow{3}{*}{6}
& NWM      & 5 & 1 & 0/6 \\
& & OmniVLA  & \textbf{0} & 2 & 4/6 \\
& & \methodname{} & \textbf{0} & \textbf{0} & \textbf{6/6} \\
\midrule
\multirow{3}{*}{Meeting} & \multirow{3}{*}{6}
& NWM      & 3 & 2 & 1/6 \\
& & OmniVLA  & \textbf{0} & 3 & 3/6 \\
& & \methodname{} & 2 & \textbf{0} & \textbf{4/6} \\
\midrule
\multirow{3}{*}{Hallway} & \multirow{3}{*}{4}
& NWM      & 2 & \textbf{0} & 2/4 \\
& & OmniVLA  & \textbf{1} & \textbf{0} & 3/4 \\
& & \methodname{} & \textbf{1} & \textbf{0} & 3/4 \\
\midrule
\multirow{3}{*}{All} & \multirow{3}{*}{24}
& NWM      & 13 & 7 & 4/24 \\
& & OmniVLA  & \textbf{3} & 7 & 14/24 \\
& & \methodname{} & \textbf{3} & \textbf{2} & \textbf{19/24} \\
\bottomrule
\end{tabular}
\end{table}

\paragraph{Qualitative Differences across Methods.}
The aggregate row of Table~\ref{tab:sup_failure_modes} shows that the three methods fail in qualitatively different ways. NWM fails mainly by drift ($13$ drift vs.\ $7$ collision), OmniVLA~\cite{omnivla} mainly by collision ($3$ vs.\ $7$), while \methodname{} has the fewest failures overall ($3$ vs.\ $2$). This pattern is consistent with each method's control structure: NWM's open-loop CEM rollout accumulates heading errors, OmniVLA's short direct trajectories reduce drift but provide limited obstacle-aware lookahead, and \methodname{} combines closed-loop replanning with joint action--future--value supervision. As a result, \methodname{} achieves the lowest count in \emph{both} drift and collision failures.

\subsection{Additional Qualitative Results}

\subsubsection{Real-Robot Future-View Predictions}
\label{sup:qual_realrobot_fpv}

\begin{figure}[h]
\centering
\includegraphics[keepaspectratio, width=\textwidth]{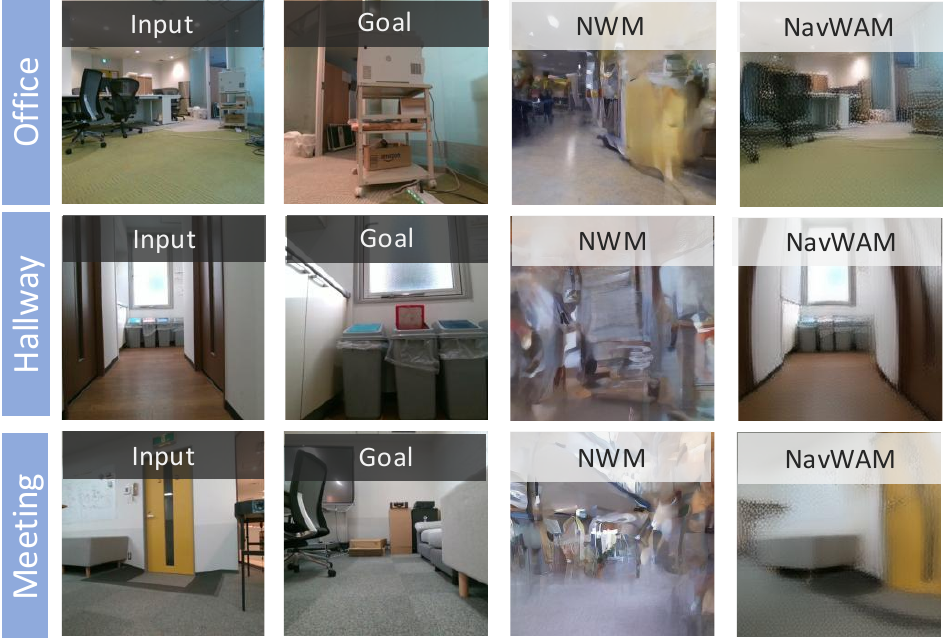}
\vspace{-0.3cm}
\caption{
\textbf{Real-robot future-view predictions: NWM vs.\ \methodname{}.}
Three scenes (office, hallway, meeting room). For each scene, from left to right: input observation, goal image, NWM's prediction at step~8, \methodname{}'s prediction at step~4.
}

\label{fig:sup_qual_real}
\end{figure}

Figure~\ref{fig:sup_qual_real} shows future-view predictions from NWM and \methodname{} on three real-robot scenes: office, hallway, and meeting room. For each scene, we show the input observation, the goal image, NWM's predicted future at $t{=}8$, and \methodname{}'s predicted future at $t{=}4$, from left to right. NWM fails to produce a consistent future view in all three scenes, indicating that its zero-shot future prediction collapses on the real-robot domain. In contrast, \methodname{} produces visually coherent futures aligned with the goal image, correctly anticipating whether the agent should move forward, move forward while turning right, or turn left according to the spatial relation implied by the goal.

\subsubsection{Real-Robot Value Trajectory}
\label{sup:qual_value}

\begin{figure}[h]
\centering
\includegraphics[keepaspectratio, width=\textwidth]{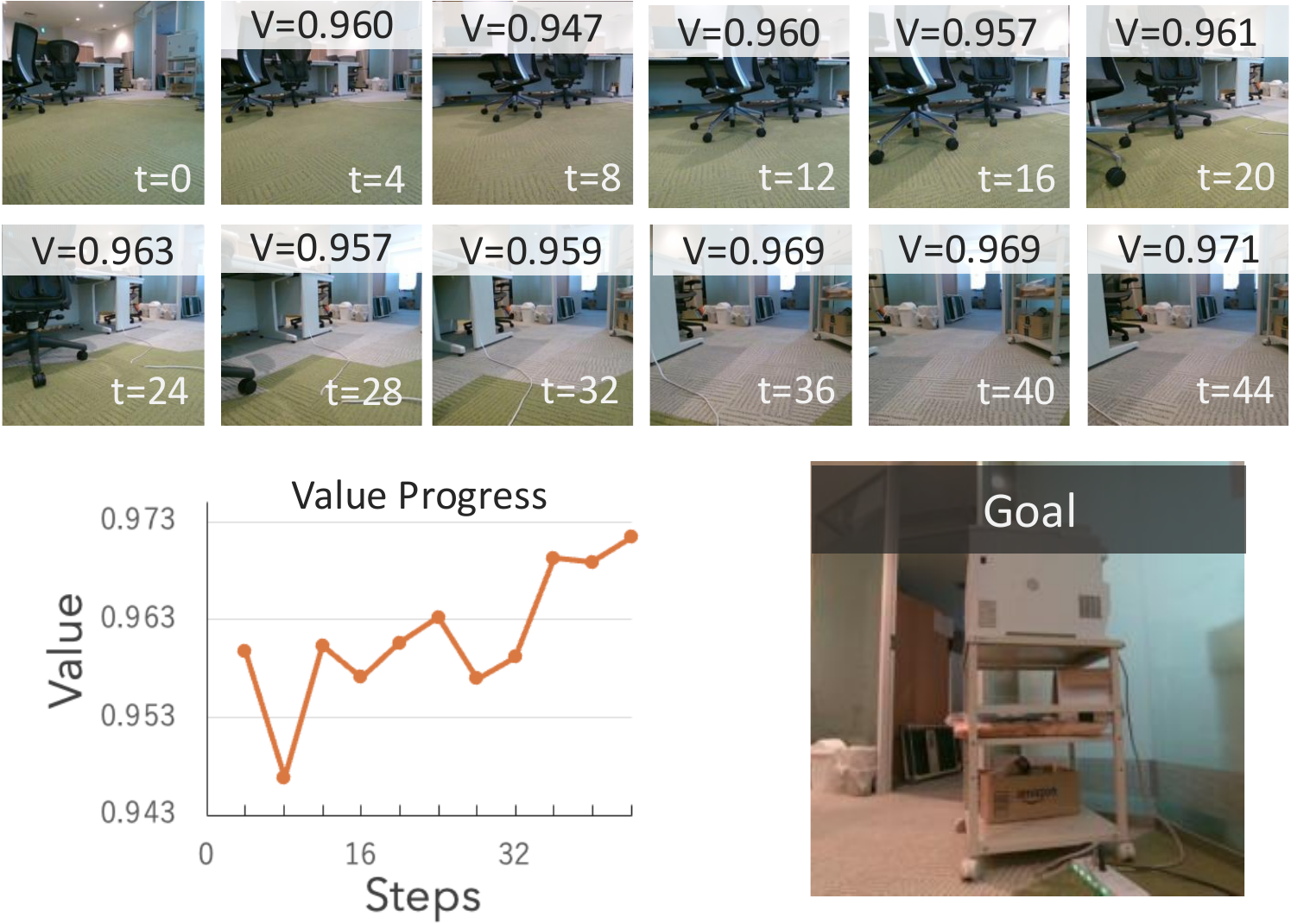}
\vspace{-0.3cm}
\caption{\textbf{Predicted value over a real-robot rollout.} Predicted scalar value at each step together with the FPV at selected steps. Single-pass execution.}

\label{fig:value}
\end{figure}

In addition to the action chunk and future image, \methodname{} predicts a scalar goal-progress value (Eq.~\ref{eq:sup_value}). Figure~\ref{fig:value} shows the predicted value over a single-pass real-robot rollout (no Best-of-$N$ sampling) together with the FPV at selected steps. At $t{=}0$ and $t{=}4$ the goal object (a copier) is visible on the right of the FPV and the predicted value sits near the top of its dynamic range; as the robot follows a gentle curve and the copier leaves the view by $t{=}8$, the predicted value drops sharply. From $t{=}12$ onwards the robot reorients toward the goal region and the predicted value partially recovers through $t{=}44$. Two properties make this trajectory more than time-of-rollout noise. \emph{(i)~Anchoring to goal visibility:} The largest excursion ($t{=}8$) coincides with the goal object leaving the FPV, so the value is genuinely conditioned on the goal-image context rather than tracking a generic rollout-time signal. \emph{(ii)~Closed-loop self-consistency:} The recovery from $t{=}12$ onwards is produced by the same model weights re-evaluating each chunk on the new observations that follow the previous chunk's execution, so the upward trend is not an artefact of teacher forcing or external annotation; it is the head's own assessment that the post-execution observations imply higher goal progress. We note that the value is calibrated to full-trajectory distances via Eq.~\ref{eq:sup_value}, so its absolute dynamic range over short real-robot rollouts is compressed; the dynamics here should be read qualitatively rather than as a calibrated progress probability.

~\section{Limitations and Broader Impact}
\label{sup:limitations}

The main paper notes limitations briefly. We expand here on the regimes in which \methodname{} is expected to fail and the broader implications of releasing a navigation policy of this type.

\subsection{Failure Mechanisms}
\label{sup:fail_regimes}

\begin{figure}[h]
\centering
\includegraphics[keepaspectratio, width=\textwidth]{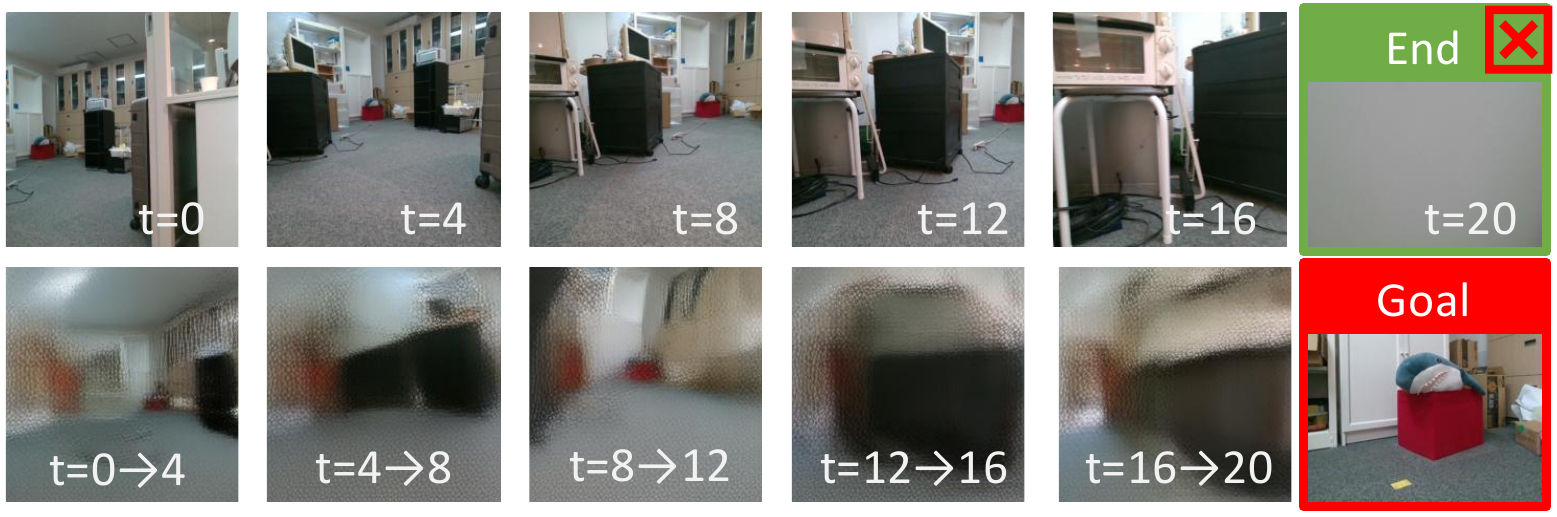}
\vspace{-0.3cm}
\caption{\textbf{Representative real-world failure of \methodname{}.} Egocentric FPV and predicted future image at each step.}
\label{fig:sup_qual_diablo_failure}
\end{figure}

\paragraph{Observed Failure.}
Figure~\ref{fig:sup_qual_diablo_failure} shows a representative failure of \methodname{}, including the egocentric FPV and the predicted future image at each step. Up to $t{=}8$, the predicted futures match the executed observations, and the actions remain goal-directed. At $t{=}12$, however, the robot's actual pose deviates from the prediction: the robot has over-rotated to the left, likely due to wheel friction. From this off-trajectory observation, the model predicts an inconsistent future, which the action chunk then follows. As a result, the trajectory drifts and the robot collides.
This failure highlights two compounding mechanisms. \emph{(i)~Pose drift:} Physical effects such as friction or slippage push the robot away from the predicted trajectory, so subsequent FPVs no longer match the regime on which the future predictor was trained, degrading prediction quality. \emph{(ii)~Degraded prediction near obstacles or in visually ambiguous regions:} When the robot approaches an obstacle too closely or enters a visually ambiguous region, the future-view prediction degrades, and the action chunk drifts with it.

\paragraph{Unverified Regimes.}
We do not evaluate the following two regimes and leave them to future work.
\begin{itemize}[leftmargin=1.2em,itemsep=2pt,topsep=2pt]
    \item \textbf{Dynamic obstacles.} Our real-robot deployment is limited to static scenes, so \methodname{}'s behavior in the presence of moving people or other dynamic obstacles remains unverified.
    \item \textbf{Long-horizon navigation.} Our real-robot deployment covers short- to medium-range goals within a single room or corridor. Long-horizon settings, such as multi-room or multi-floor navigation requiring many replanning chunks, remain unverified.
\end{itemize}

\subsection{Detailed Limitations}
\label{sup:detailed_limitations}

In addition to the limitations summarized in the main paper, we note the following:

\begin{itemize}[leftmargin=1.2em,itemsep=2pt,topsep=2pt]
    \item \textbf{Evaluation breadth.} The main paper focuses on image-goal navigation. Language-goal navigation, object-goal navigation, instruction-following, and embodied question answering are not evaluated.
    \item \textbf{Real-world scale.} The closed-loop deployment covers $24$ episodes across four indoor environments on a single robot platform. Larger and more diverse real-world evaluation, in particular across robot platforms with different cameras and control rates, is left for future work.
    \item \textbf{Inference cost.} \methodname{}'s single-pass cost (Supp.~\ref{sup:efficiency}) is dominated by the diffusion chain length. Reducing the chain length trades latency for accuracy along the standard diffusion curve; aggressive step reduction has not been studied in the main paper.
    \item \textbf{Value function calibration.} The value frame is calibrated to the indoor trajectory distributions seen during fine-tune (Eq.~\ref{eq:sup_value}). Its absolute values are not directly comparable across datasets with different scale.
    \item \textbf{Action and value head design.} We use direct readout from the denoised action and value frames; attaching separate MLP heads on top of these frames gave no improvement and increased complexity during development, so we do not explore this design further in the main paper.
\end{itemize}



\paragraph{Safety and Deployment.}
\methodname{} is a closed-loop visual-navigation policy that produces locally planned motion commands from egocentric observations and a goal specification. Our real-world deployment uses a Diablo platform under operator supervision, with a per-episode time budget and a hardware safety stop. We recommend the same level of supervision for any reproduction of the closed-loop results. The released checkpoints and code are intended for research use and are not safety-certified for autonomous operation around people or in safety-critical environments.


\end{document}